# Constraining Lexical Selection Across Languages Using TAGs*

Dania Egedi
Institute for Research in Cognitive Science
University of Pennsylvania
Philadelphia PA 19104-6228
egedi@linc.cis.upenn.edu

Martha Palmer
Department of Computer Science
University of Pennsylvania
Philadelphia PA 19104-6389
mpalmer@linc.cis.upenn.edu

Lexical selection in Machine Translation consists of several related components. Two that have received a lot of attention are lexical mapping from an underlying concept or lexical item, and choosing the correct subcategorization frame based on argument structure. Because most MT applications are small or relatively domain specific, a third component of lexical selection is generally overlooked - distinguishing between lexical items that are closely related conceptually. While some MT systems have proposed using a 'world knowledge' module to decide which word is more appropriate based on various pragmatic or stylistic constraints, we are interested in seeing how much we can accomplish using a combination of syntax and lexical semantics. By using separate ontologies for each language implemented in FB-LTAGs, we are able to elegantly model the more specific and language dependent syntactic and semantic distinctions necessary to further filter the choice of the lexical item.

## 1 Introduction

One of the primary tasks for machine translation is lexical selection - selecting the target lexical item that most closely matches the source lexical item being translated. For transfer based approaches such as Transtar [17] and Geta [15], each separate lexeme in the source language must be paired with a corresponding lexeme in the target language in a set of bilingual dictionaries. An alternative[1] is the interlingua approach, such as Princitran [3] or Translator [8], in which the source verb is mapped to a canonical semantic representation which is shared by all target languages. The elements of the semantic representation select the lexical choice in each target language.

There are several components of lexical selection. Two that have received a lot of attention are lexical mappings from an underlying concept or lexical item, and choosing the correct subcategorization frame based on argument structure. Because most MT applications are small or relatively domain specific, a third component of lexical selection is generally overlooked - distinguishing between lexical items that are closely related conceptually. There can be many shades of distinction between the meaning of a lexical item in one language and its counterpart in another language [11]. These distinctions are sometimes critical to selecting the correct lexical item in the target language. The question then arises, in both transfer and interlingua based systems, of how and where to capture these distinctions. While some MT systems have relegated this task to a 'world knowledge' or 'pragmatics' module [2, 14], we are interested in seeing how much we can accomplish using a combination of syntax and lexical semantics. In this paper, we outline a proposal to capture these distinctions based on separate ontologies for each individual language. Our method is applicable to both transfer and interlingua based approaches, and provides a more elegant solution than exhaustive enumeration and a more local solution then reliance on 'world knowledge' modules. This method has been partially implemented in FB-LTAGs [4, 12, 16], whose feature-based, lexicalized approach provides an advantageous environment for modelling the more specific and language dependent syntactic and semantic distinctions necessary to further filter the choice of the lexical item.

## 2 Defining the Problem

The essence of the problem that we are trying to solve involves lexical constraints that are critical for one language but non-existant or completely different in another. A classic example of this is the Japanese *wear* example.

(1) *kare wa boushi wo kaburu.*
    he      hat      wear
    *He wears a hat.*

(2) *kare wa kutsushita wo haku.*
    he      socks         wear

---

*We would like to thank Aravind Joshi, Sadao Kurohashi, and Zhibiao Wu for their helpful input.

[1] These are the two ends of the spectrum, and many systems now take a hybrid approach. Since the purpose of this paper is to highlight a area of MT usually ignored, and to propose a non-theory specific solution, we will not give an overview of all types of MT systems. We do limit our initial comments to non-statistical MT methods, as we do not believe that our method would be useful to purely statistical systems.

*He wears a pair of socks.*

Sentences (1) and (2) highlight a situation in which one language (Japanese) distinguishes several senses of a concept /WEAR/ that has only one sense in another language (English). In Japanese, *kaburu* selects for items worn on the head, such as *hats*, while *haku* selects for items such as *socks*. English *wear* does not make this lexical distinction.

A similar problem occurs when translating English *break* into Chinese. The semantic features used in selecting the correct verb construction in Chinese (such as the initial shape of the object, choice of instrument) are not all used in selecting English verb senses. This causes difficulties for a large-scale transfer based system such as TRANSTAR, a commercial broad coverage English/Chinese MT system developed in Beijing. When this system is applied to sentences from the Brown corpus that contain *break*, an accuracy rate of less than 20% is achieved, even after ruling out idiomatic uses and problems with parsing [17]. The primary reason is that in English *break* can be thought of as a very general verb indicating an entire set of breaking events which can be distinguished by the resulting state of the object being broken. *Shatter*, *snap*, *split*, etc. can all be seen as more specialized versions of the general breaking event. Chinese has no equivalent verb for indicating this class of breaking events, and each usage of *break* has to be mapped onto a more specialized lexical item. Even the English specializations of a breaking event do not cover all of the different ways in which Chinese can semantically distinguish between breaking events. The end result is that lexical selection from English to Chinese is often predicated on the existence of semantic features that are completely irrelevant to English.

This is not a problem that is unique to English and Chinese or Japanese. In looking for cross-linguistic semantic universals for *break* and other semantically similar verbs, Pye found that there were as many different semantic classification schemes as there were languages being investigated [11]. The solution to this problem is elusive enough when considering two particular languages. It must be recognized that a typical transfer-based approach requires a direct mapping from each distinct verb sense to its corresponding lexical item in the target language, and must therefore specify all of the semantic features relevant to both languages. The interlingua approach has a similar difficulty, since it must define an interlingua that can take into account all of the semantic features for both languages. When one begins to consider the problem from the perspective of several languages, this technique quickly becomes impractical. The direct mapping approach becomes cumbersome, unwieldy, and extremely tedious to build, since it means reanalyzing the semantic features of each language according to every language that it is being paired with. For the interlingua, a vast, language universal ontology must be built that incorporates every semantic feature for every language in an organized fashion. That means that not only do correspondences have to be found between individual lexical items, but also between the classification schemes by which each language structures its concepts. While there has been a lot of promising recent work on the problem of verb classifcation, but it is not clear that it supports the notion of a readily accessible language universal ontology. For instance, Levin [6] has shown that there is a correspondence between lexical-semantic verb classes and syntactic structure for English and there has been speculation that these verb classes should extend to other languages since they are based on cross-linguistic semantic concepts. Mitamura, however, has determined a classification for Japanese verbs that shows very little correspondence to Levin's classes [7]. The EDR project, an enormous effort (over 200,000 words) to build an English-Japanese bilingual dictionary based on a joint conceptual classification has found a conceptual overlap between the two languages of only about 10% [18]. Another large ongoing effort in France has also been looking at generalizations about verb classes in French that can be made based on allowable syntactic transformations. This work is currently being extended to several other languages, but each language is being done independently, from the ground up, with very little sharing of classification schemes [5]. None of this rules out the possibility of semantic universals, or large areas of conceptual overlap between languages, but it does highlight the extreme individuality of each language, and the overwhelming task that lies in front of anyone trying to merge language-specific conceptualizations.

## 3 Proposed Model

We believe that the most practical approach is to assume that each language will require its own conceptual ontology with a distinct set of semantic features. Many of the concepts in the lexical semantic ontologies may be shared among languages, but languages may choose to structure the concepts differently. With this in mind, we suggest an approach to translation that does not always attempt to directly map a specific verb sense in the source language to another specific sense in the target language. Rather, it begins with a more coarse-grained lexical translation process, which merely attempts to focus on a particular set of translation candidates in the source language. These candidates will be further narrowed down by a language specific lexical selection process which examines the semantic features associated with the instantiated verb arguments and determines the best fit. Therefore, in many cases, the detailed merging of language specific semantic features associated with the source sense to the target sense can simply be avoided. Rather than one-to-one mappings between lexical items, the dictionary would map between sets of lexical items[2]. For the transfer approach,

---
[2]This is similar in motivation to the interlingua approach, where the goal is to capture semantic similarities by associating several lexical items with the same primitive concept. In the same way, we are grouping semantically similar lexical items

one consequence of making the semantic structures local is a much broader concept of the bilingual dictionary. For instance, English *break* maps to a set of Chinese verbs such as *da sui* (break into many pieces), *da puneig* (break continuity), *da po* (break into irregularly shaped pieces). Correspondingly, *da sui* would map to a set of English verbs such as *break, shatter, crumble*. The final selection of the actual lexical item will be made in the target language based on the semantic features associated with the prospective arguments. This type of of approach can be used to match a lexical item to another lexical item, or it could also be used to match a lexical item to an 'deep semantic' representation, such as an interlingua. As such, it could be utilized in either a transfer or interlingua based system. One of the advantages of this approach is that the same self-contained, language-specific representation that is normally used for any form of analysis or generation becomes very applicable to machine translation [9]. More importantly, it is not necessary that the languages being translated have the same underlying verb classes, since the semantic structure is local to each language. However, we cannot entirely avoid the issue of finding the conceptual links between language-specific classification schemes. We are still left with the problem, given the different classification schemes, of associating appropriate classes of lexical items in a target language with the most closely corresponding class in the source language. Since we have just argued that there will never be exactly corresponding classes in any two languages, this is clearly still a difficult issue. However, we do not have to try to force the different classification schemes into a single interlingua. It might be that the most useful method for taking advantage of our approach would be in a hybrid system that uses a direct transfer method in certain circumstances, and a more general, classification-correspondence approach in other circumstances.

## 4 Implementation

We have begun to implement this model in a variant of the Synchronous TAGs formalism, a Lexicalized TAG suitable for machine translation [13, 1], which has been augmented to handle feature-based unification. This particular formalism has a number of advantages for our approach. First, it is lexicalized, which makes it easier to specify the lexically specific semantic information in a syntactic context. This is important in languages such as English where the semantics can have syntactic consequences [6]. Second, it is feature-based, which provides a convenient notation and mechanism in which to specify the selectional restrictions. Third, the extended domain of locality provided by the tree structures allows lexical items to easily place constraints on other lexical items in the same frame. The disadvantages of FB-LTAGs include difficulty in specifying complex feature hierarchies, and a unification system that would not be able to take advantage of class inclusion.

Although Synchronous TAGs can also be used with an interlingua [14] we chose to start with a transfer-based approach between the languages Chinese and English. We will work through an simple example to show how Synchronous FB-LTAGs handle this method. Semantic constraints are specified in the usual method for each language. The semantic characteristics of a lexical item (or each sense of a lexical item) are instantiated as features in the syntactic lexicon. Lexical items may also specify constraints on semantic features of other lexical items available in its syntactic frame (i.e. local to its tree). At parse time, of course, the features and feature constraints must unify. This is done independantly for each language.

Our syntactic lexicon lists, among others, 5 Chinese expressions for the concept *break*:

- *da sui* (hit into many pieces)
- *da duan* (break into line segments)
- *da po* (hit irregular)
- *da hui* (make nonfunctional)
- *da puneig* (discontinue a journey or song)

The lexical item for each Chinese verb specifies in its features what semantic restrictions it places on its object[3]. Each noun also specifies its semantic categories, at the granularity that is necessary for this particular language. For instance, the Chinese verb *da sui* takes an object that is a physical object and is brittle, while the verb *da puneig* takes a continuous abstract object, as illustrated in Figure 1. The noun *huapin* (vase) is, among other things, a physical, irregularly shaped, brittle object, and the noun *lucheng* (journey) is a continuous, abstract object. The corresponding noun phrase trees are shown in Figure 2.

When translating the English sentence *John broke the vase*, the lexical transfer for *break* maps from the semantic concept /BREAK/ in English to the semantic concept /BREAK/ in Chinese. This includes all 5 Chinese expressions listed above. Thus a number of Chinese translations are initially generated, but the semantic feature constraints imposed by most of the verbs will cause the sentence to fail. For instance, *da puneig* would not be able to unify with the noun *huapin*. The sentence *John broke the vase* would then necessarily be translated as *Ji-Yong da sui huapin*, while *John broke the journey* would be translated as *Ji-Yong da puneig lucheng*. We were able to correctly translate the the English sentence without specifying semantic

---
together, but we are retaining the complete semantic representations with selection restrictions for the individual lexical items. The group "class" or "concept" is not a substitute for the individual semantic representation, but an enhancement.

[3]The same is true on the English side where we differentiate between different senses of the lexical item *break*, which are distinguished by the object of the clause, i.e. functional break vs physical break [10]. These senses have different syntactic behaviors in English. Critically, though, the distinctions necessary for English are not forced onto the Chinese breaking verbs (or vice versa).

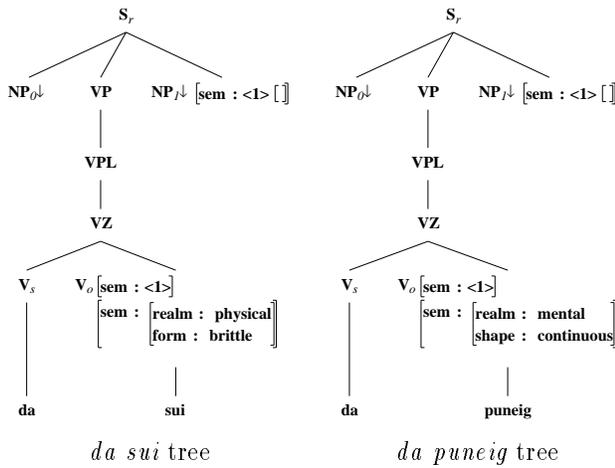

Figure 1: Two trees corresponding to English *break*

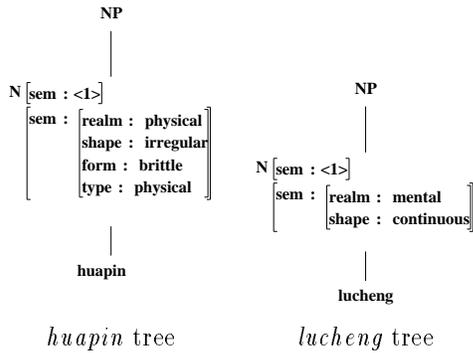

Figure 2: NP trees for Chinese *vase* and *journey*

information in the English syntactic lexicon that was critical to the correct lexical selection in Chinese.

## 5 Future Work and Conclusion

This work is initial work on a problem of Machine Translation that has often been ignored or relegated to 'pragmatics' or 'world knowledge'. As such, there remains much more work to done, from extending our implementation described here to include a larger set of lexical items, to working on semantic ontologies for the languages that we are interested in, to questions such as how much and what kind of information is really language specific. Unless we are claiming that no features need to be shared between language translation pairs, which we are not, a decision must still be made about what information should be transferred between the languages. A related question arises for interlingua approaches - what information should be included in the underlying semantic representation. It is not at all clear to us where that line should be drawn.